\documentclass[%
 aps,pra,10pt,
 amsmath,amssymb,showkeys
]{revtex4-2}

\usepackage{epsfig,pstricks}
\usepackage{graphicx}
\usepackage{xcolor}
\usepackage{booktabs}
\usepackage{enumitem}
\usepackage{physics}
\usepackage{mathtools}
\usepackage{float}
\usepackage{overpic}  

\begin{document}

\title{Image classification via a quantum-inspired strategy involving a mixture of experts}

\author{Kumari Jyoti}
\email{kumarijyoti@iisc.ac.in}
\affiliation{%
Centre for High Energy Physics, Indian Institute of Science, Bangalore 560012
}%

\author{Rohith Babu}
\email{rohithbabu@iisc.ac.in}
\affiliation{%
Centre for High Energy Physics, Indian Institute of Science, Bangalore 560012
}%

\author{Apoorva D. Patel}
\email{adpatel@iisc.ac.in}
\affiliation{%
Centre for High Energy Physics, Indian Institute of Science, Bangalore 560012
}%

\date{\today}

\begin{abstract}
Pattern recognition problems arise in a variety of physical image processing situations, and convolutional neural networks are a popular scheme for the required feature extraction and classification tasks. The classical networks use diffusion-based smearing and block-wise pooling to downsample the image data and capture important structural features. In this work, we propose and demonstrate a more efficient quantum-inspired strategy involving a mixture of experts. It is a hybrid classical-quantum framework. The quantum part consists of amplitude encoding of the images, convolution using local unitary operations, multiple experts processing the same image with different parameters, and feature extraction using quantum stabiliser codes. The classical part then jointly processes the features extracted by different experts using a standard fully connected neural network for image class prediction. Using MNIST and Fashion-MNIST datasets as benchmarks, we demonstrate that the joint expert analysis outperforms the individual expert one, as well as reduces the failure rate of image class prediction by around a factor of two. The overhead of our quantum-inspired strategy is only moderate on GPU workstations, which makes our proposal a practical alternative to existing classical schemes. We also point out how the quantum part of our framework can be executed on a quantum processor.
\end{abstract} 


\maketitle


\section{Introduction}\label{sec:intro}

Rapid technological developments have made various types of sensors and detectors affordable and convenient. They collect huge amounts of data which need to be analysed to make decisions. Often there is no time or space to store the data; interesting features must be extracted quickly and the rest discarded. Examples range from astronomy, imagery and surveillance to internet traffic and collider physics. Machine learning models are used for pattern recognition analysis of such big data problems, without direct human intervention.

Supervised learning is a basic machine learning problem, where the models are trained using data with known pattern labels, and then used to classify unlabeled test data. In the standard form, the data is first embedded in a high dimensional feature space, and then separated into disjoint classes with binary labels using linear algebra operations. The common technique processes the high-dimensional data using Convolutional Neural Networks (CNNs) \cite{krizhevsky2012,hinton2015} for feature extraction and spatial aggregation. Mechanisms such as diffusion and pooling utilise continuous neighborhood dynamics and heat-kernel approximations to reduce the data while capturing multi-scale features. While mathematically robust, these manipulations are inherently lossy. Nevertheless they are imperative, because handling data with a large number of features is computationally prohibitive and only a small number of features are important for extracting the relevant information such as the class labels. The central aim then is to find a clever way to reduce the number of features, such that as much relevant information as possible can be extracted from the reduced data.

In recent years, quantum algorithms have been developed for various information processing tasks. Many attempts are being made to see how incorporating quantum logic operations in machine learning strategies can enhance their efficiency. It is becoming clear that significant quantum advantage can be obtained only when quantum data are provided as input to the algorithms \cite{bermejo2026,bowles2024,QMLrev}. Such a scenario can be a genuine quantum experiment whose output is fed to a quantum processor, without any measurement that would collapse the output to a classical form \cite{huang2022}.

In this work, we consider a more general structure for ``quantum data", where both magnitude and phase information can be extracted from the same data. Such situations occur when signals from a wave source contain the data of interest, and multiple detectors are used to study the source properties from different perspectives. A motivating analogy comes from how two eyes allow us to judge the distance of the source by parallax and two ears allow us to judge the direction of the source by detection time difference---properties that cannot be obtained with only one eye or one ear. We incorporate this strategy in our work, by encoding images in superposition states and then letting multiple experts classify them from different viewpoints (these viewpoints are taken to be initial values of the variational parameters that are tuned to classify the data). This defines our framework for a Quantum Mixture of Experts (QMoE) analysis, which has been implemented by others in a somewhat different manner \cite{MoEref1,MoEref2}.

We replace parts of the CNN technique by quantum subroutines to construct a hybrid image classification algorithm, where classical diffusion is replaced by unitary smearing and block-wise pooling is substituted by a quantum stabiliser code, while the binary classification task is carried out by a standard fully connected neural network. We then compare the classification performance of the quantum subroutines with their classical versions. We also check for the advantage offered by multiple experts acting jointly versus when they act independently. Our tests are carried out for both the publicly available MNIST \cite{lecun2010mnist} and Fashion-MNIST \cite{xiao2017fashion} datasets. The MNIST database consists of various handwritten forms of 0-9 digits forming 10 classes, while the Fashion-MNIST database consists of pictures of various clothing items also forming 10 classes.

Our detailed algorithmic methodology is presented in Section 2. Our results are discussed in Section 3, and we end with conclusions in Section 4. All our numerical analysis is carried out in a quantum-inspired setting, with PyTorch programming on GPU workstations. In the Appendix, we describe how our quantum subroutines can be implemented on a quantum processor.

\section{Methodology}

Our algorithmic implementation has several components: State encoding, convolutional smearing,
introduction of multiple experts, block-wise pooling, and classification using a fully connected neural network. We describe these components in turn.

\subsection{State Encoding}
Both the MNIST and Fashion-MNIST datasets are grey-scale 28$\times$28 pixel images. We pad them with blank pixels to 32$\times$32 size, for easy binary processing. Both $x$ and $y$ coordinates of the images are then labeled using 5 bits each.

We embed the images into quantum states using amplitude encoding. We first convert each grey-scale image into a black-and-white one with a thresholding step, such that the value for each pixel $c_{x,y}=$+1 or -1. We then construct the highly entangled superposition state:
\begin{equation}
    |\psi\rangle = \frac{1}{32} \sum_{x,y\in\{0,1\}^5} c_{x,y} |x_0 x_1 x_2 x_3 x_4,y_0 y_1 y_2 y_3 y_4\rangle .
    \label{eq:amp_encoding}
\end{equation}
Such an amplitude encoding of the classical pixel values is exhaustive at the beginning. But once it is done, all the subsequent operations are carried out on the encoded 10 qubit states, with an exponential reduction of the space resources.

\subsection{Convolutional Smearing}
\label{sec:unitary_convol}
The convolutional smearing mixes pixel value for each image with those of its neighbours. The classical diffusion process uses the discrete Laplacian operator for this purpose. We implement that iteratively, by repeating nearest neighbour mixing at each iteration for a number of iterations. Denoting the pixel values at time $t$ by $X_{i,j}^t$,
the transformation is:
\begin{equation}
    X_{i,j}^{t+1} = (1-\alpha)X_{i,j}^t + \frac{\alpha}{4} (X_{i-1,j}^t + X_{i+1,j}^t + X_{i,j-1}^t + X_{i,j+1}^t) .
\end{equation}
The parameter $\alpha$ controls the diffusion speed, and the transformation is repeated for $T$ time steps. This update can be easily carried out with a checkerboard partition of the square pixel grid that we have. Roughly, $\alpha T$ represents the total diffusion depth, and $\alpha \in [0,1]$ is needed for the stability of diffusion. For finite $T$ and checkerboard updates, the range of $\alpha$ can be extended somewhat to optimise smearing. Such a smoothing of the pixel values degrades fine-grained structural details, transforming the image towards a blurred low-resolution spatial distribution. The blurred images can be good enough to discern shapes, clusters and global patterns. 

\begin{figure}[ht]
    \centering
    \epsfxsize=7cm \epsffile{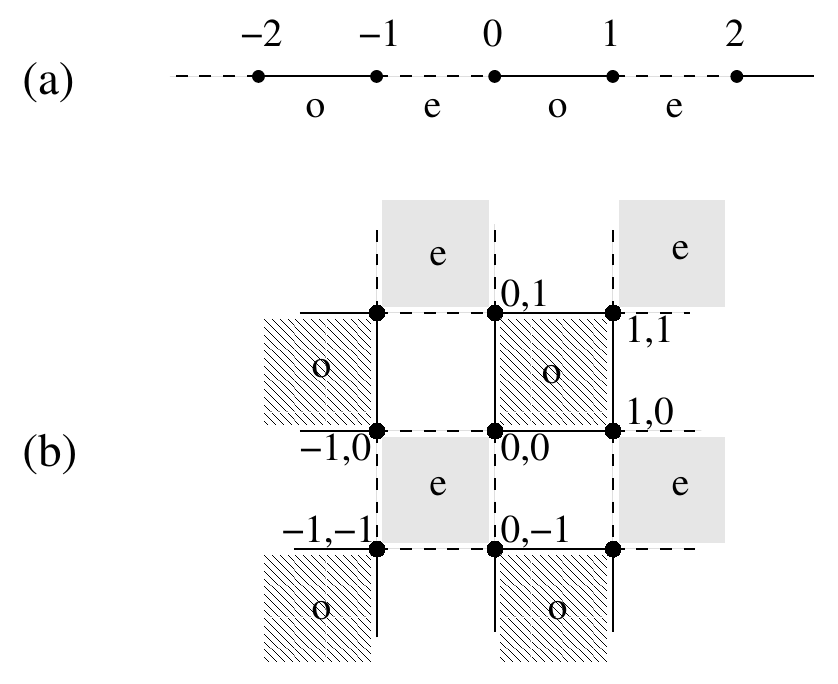}
    \caption{Bipartite splitting of hypercubic lattice structures: (a) One dimension, (b) Two dimensions}
    \label{fig:BipartiteSplit}
\end{figure}

Instead, we process the quantum image states by a non-lossy unitary smearing, using the transformation described in \cite{patel2005}. Local unitary smearing can be carried out for a hypercubic structure in any dimension, by a bipartite splitting of the data into even and odd subsets, as illustrated in Fig. \ref{fig:BipartiteSplit}.

The two-dimensional image is organised in a checkerboard pattern, and the two subsets (labeled even and odd) are evolved in an alternate manner. Each time step implements
\begin{equation}
    |\psi(t+1)\rangle = U_e U_o|\psi(t)\rangle .
\end{equation}
The operators $U_{o(e)}$ unitarily mix the quantum amplitudes at the four corners of checkerboard elementary squares, according to
\begin{equation}
    U_{o(e)} = cI - is\sqrt{2}H_{o(e)} ~,~~ |c|^2 + |s|^2 = 1 .
    \label{eq:unitary_blocks}
\end{equation}
The $4\times4$ Hamiltonian blocks are:
\begin{eqnarray}
    H_o^B &=& -\frac{i}{2} \begin{pmatrix} 0 & -1 & -1 &  0 \\
                                           1 &  0 &  0 &  1 \\
                                           1 &  0 &  0 & -1 \\
                                           0 & -1 &  1 &  0 \end{pmatrix}
                         \begin{matrix} 00 \\ 10 \\ 01 \\ 11 \end{matrix} \\
          &=& -\frac{1}{2} (I \otimes \sigma_2 + \sigma_2 \otimes \sigma_3) ,
\end{eqnarray}
where the column on the right denotes the vertices of the elementary square on which $H_o^B$ operates, and $\sigma_i$ are the Pauli matrices. Similarly, $H_e^B = -H_o^B$ when operating on the elementary square with vertices \{00,-10,0-1,11\}. The mixing parameters $c$ and $s$ can be tuned to achieve suitable smearing of the image, and the total smearing depth is roughly $sT$. More generally, the parameter $s$ can be chosen to be different for different time steps. Note that we have adopted the convention such that all numerical computations need only real numbers (and not complex numbers).

\subsection{Multiple Experts}
We identify the experts by the parameters they use for the classification problem. These parameters appear in the convolutional smearing part of the algorithm, while the block-wise pooling part is essentially deterministic. In the classical case, the relevant parameter is $\alpha$, and the optimisation of classification fidelity amounts to finding the optimal value for $\alpha$.

In the quantum case, the experts are labeled by their parameter values for $s$. There are two strategies available. We can optimise the classification fidelity for each expert independently, and make the final classification prediction by a majority vote. Alternately, we can optimise the classification fidelity by joint decisions made by all the experts together. Our emphasis in this work is on showing that joint decisions by all the experts provides better classification predictions compared to the case when the experts act independently.

\subsection{Block-wise Pooling}
\begin{figure}[ht]
\begin{center}
{\setlength{\unitlength}{1mm}
\begin{picture}(30,25)
\thicklines
\put(5,20){\line(1,0){20}}
\put(5,10){\line(1,0){20}}
\put(5,0){\line(1,0){20}}
\put(5,0){\line(0,1){20}}
\put(15,0){\line(0,1){20}}
\put(25,0){\line(0,1){20}}
\put(5,0){\circle*{2}}
\put(15,0){\circle*{2}}
\put(25,0){\circle*{2}}
\put(5,10){\circle*{2}}
\put(15,10){\circle*{2}}
\put(25,10){\circle*{2}}
\put(5,20){\circle*{2}}
\put(15,20){\circle*{2}}
\put(25,20){\circle*{2}}
\put(16,11){1.0}
\put(16,21){0.5}
\put(16,1){0.5}
\put(0,11){0.5}
\put(26,11){0.5}
\put(-2,1){0.25}
\put(26,1){0.25}
\put(-2,21){0.25}
\put(26,21){0.25}
\end{picture}
}
\end{center}
\caption{Pixel weights for classical pooling using a 3$\times$3 window}
\label{fig:AveragingWeights}
\end{figure}
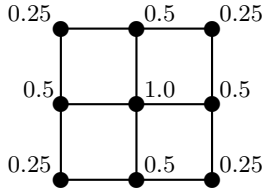

Block-wise pooling is an irreversible process that reduces the number of image pixels, while retaining the features crucial for image classification. The decrease in number of image pixels significantly cuts down the subsequent work to be carried out by the image classifier. In the classical case, we downsize the images thrice in steps of 2 along both axes, reducing the pixels first from 32$\times$32 to 16$\times$16, then from 16$\times$16 to 8$\times$8, and then from 8$\times$8 to 4$\times$4. Each downsizing step of 2 is accomplished by combining the central pixel value with those of its nearest and next-to-nearest neighbours in a 3$\times$3 window, with weights assigned as in Fig. \ref{fig:AveragingWeights}. The weights are chosen so as to give the same overall weight to each pixel, and the 3$\times$3 window slides over the whole image in steps of 2.

\noindent The final 4$\times$4 pixel values are passed on to the image classifier as 16 distinct features.

In the quantum case, we exploit the fact that error correcting codes partition the state space into non-overlapping disjoint subsets. These subsets are identified by their distinct syndromes, and the syndromes can be used as the features to be passed on to the image classifier. In this work, the pixel coordinates are labeled using 5 qubits along each axis, and so we use the perfect nondegenerate [[5,1,3]] stabiliser code to extract the syndromes for each axis \cite{preskill_chap7}. Using the four commuting syndrome generators $XZZXI$, $IXZZX$, $XIXZZ$ and $ZXIXZ$ for each axis, we extract the 4$\times$4=16 distinct features (they are the expectation values of the syndrome generators in the state $|\psi\rangle$) that are passed on to the image classifier. Since the stabiliser generators are independent of the images, they can be precomputed and stored in lookup arrays. Then the syndrome extraction is considerably simplified by observing that when the generators are expressed as 32$\times$32 matrices, there is only one nonzero element (which is either +1 or -1) in each row and each column. Note that the [[5,1,3]] stabiliser code does not distinguish between the two highly entangled states corresponding to the encoded logical qubit, and we ignore the associated feature.

\subsection{Classification}
We use the popular strategy of a fully connected neural network with multiple layers (FCL) to classify the images by processing their extracted features. Such fully connected neural networks have a large number of parameters, which frequently lead to problems of barren plateaus and local minima, because many parameters are not relevant to the classification problem. The significant reduction in the number of features, resulting from the convolution and pooling processes, simplify the classification problem by making the required FCL smaller. 

Both MNIST and Fashion-MNIST datasets have 10 distinct classes, digits and shapes respectively. We perform a binary classification analysis, i.e. whether the test image belongs to a particular class or not, and repeat that for each of the 10 classes. In both classical and quantum cases, we have 16 distinct features extracted from the images that are fed to the FCL (compared to the 784 pixels of the original 28$\times$28 images).

We experimented with the PyTorch implementation of FCL to find a reasonable setting. We chose to cascade the pooled features through a funnel-shaped $512 \to 256 \to 128 \to 10$ neural network configuration, ending with the output classes. This choice was stabilised via the batch normalization scaled by 10 and the dropout rate set to 0.2. Also, the classical baseline was evaluated for 10 different diffusion time steps, $T$=1 to 10, to benchmark the inverse relationship between diffusion and classification accuracy.

Optimisation in the FCL used loss gradient and the {\tt CrossEntropyLoss(label\_smoothing=\ldots)} function from PyTorch. For the training data, the mismatch between predicted and known class labels gives an estimate of the loss gradient for the parameters, which is then backpropagated to the convolution stage. In this parameter training process, we applied the Adam optimizer \cite {kingma2014adam} with a learning rate of 0.002.

\section{Results and Discussion}

Both MNIST and Fashion-MNIST databases have 60000 training images and 10000 test images. We carried out their classification analysis using PyTorch tensor broadcasting and vectorised linear algebra operations to get maximum computational speedup on NVIDIA H200 GPUs. In this section, we describe and discuss our results.

\subsection{Convolutional Smearing}
We smear the image pixels with multiple rounds of convolution operations. The rate of smearing is controlled by the parameter $\alpha$ in the classical case and the parameter $s\equiv\sin\theta$ in the quantum case. These parameters are later optimised by FCL. To handle coordinate shifts that go outside the image boundaries, we used modulo arithmetic to wrap around to the opposite side (these wrap-arounds involve only the padded blank pixels).

\begin{figure}[h]
    \centering
    \includegraphics[width=0.6\linewidth]{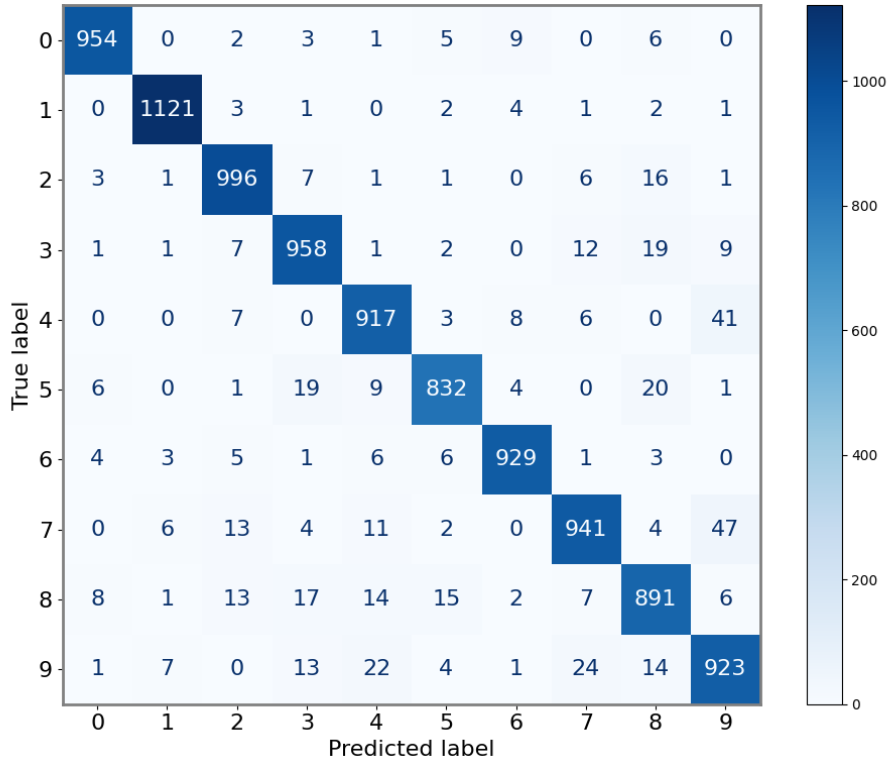} 
    \caption{Confusion matrix for the MNIST dataset in case of classical diffusion with $\alpha$=1, $T$=1.}
    \label{fig:confusion_MNIST_alpha1}
\end{figure}

In the classical case, we checked how varying the diffusion rate $\alpha$ affects the classification prediction, with $T$ fixed at 1. For the MNIST dataset, the test accuracy peaked at a 95.02\% for $\alpha=1.5$, and was worse at 94.62\% for $\alpha=1.0$ and 93.27\% for $\alpha=0.5$. For the Fashion-MNIST dataset, the test accuracy remained smaller, around 76\%. A particular example of the confusion matrix, which specifies misidentified class labels, is shown in Fig. \ref{fig:confusion_MNIST_alpha1}. Results for classification accuracy for different values of $\alpha$ are summarised in Table \ref{tab:diffusion_performance}. 

\begin{table*}[ht]
\centering
\caption{Classical classification performance for $T$=1 and varying diffusion rate $\alpha$}
\label{tab:diffusion_performance}
\small
\begin{tabular}{lcccc}
\toprule
\textbf{Dataset} & \textbf{Diffusion Rate ($\alpha$)} & \textbf{Training Accuracy} & \textbf{Testing Accuracy} & \textbf{Loss Value} \\
\midrule
MNIST  & 0.5 & 93.72\% & 93.27\% & 0.4654 \\
       & 1.0 & 94.53\% & 94.62\% & 0.4418 \\
       & 1.5 & 94.78\% & 95.02\% & 0.4368 \\
\midrule
Fashion-MNIST & 0.5 & 76.80\% & 76.30\% & 0.8282 \\
              & 1.0 & 77.00\% & 76.80\% & 0.8324 \\
              & 1.5 & 76.20\% & 75.90\% & 0.8499 \\
\bottomrule
\end{tabular}
\end{table*}

In the quantum case, for computational efficiency on a GPU, we bypassed the sequential ``for loops" using PyTorch tensor slicing. By fully extracting the subgrids in steps of 2, e.g. \texttt{X[:,:,0::2,0::2]}, the algorithm applies the $U_o$ and $U_e$ transformations to all 2$\times$2 squares at once, tantamount to a quantum processor taking advantage of superposition. In general, the parameter $\theta$ can be made dependent on the pixel position and the convolution layer number. We avoided these extra dependencies for $\theta$, because our tests showed that they needlessly increased the number of parameters without offering any accuracy advantage.

\begin{figure}[ht]
    \centering
    \includegraphics[width=\linewidth]{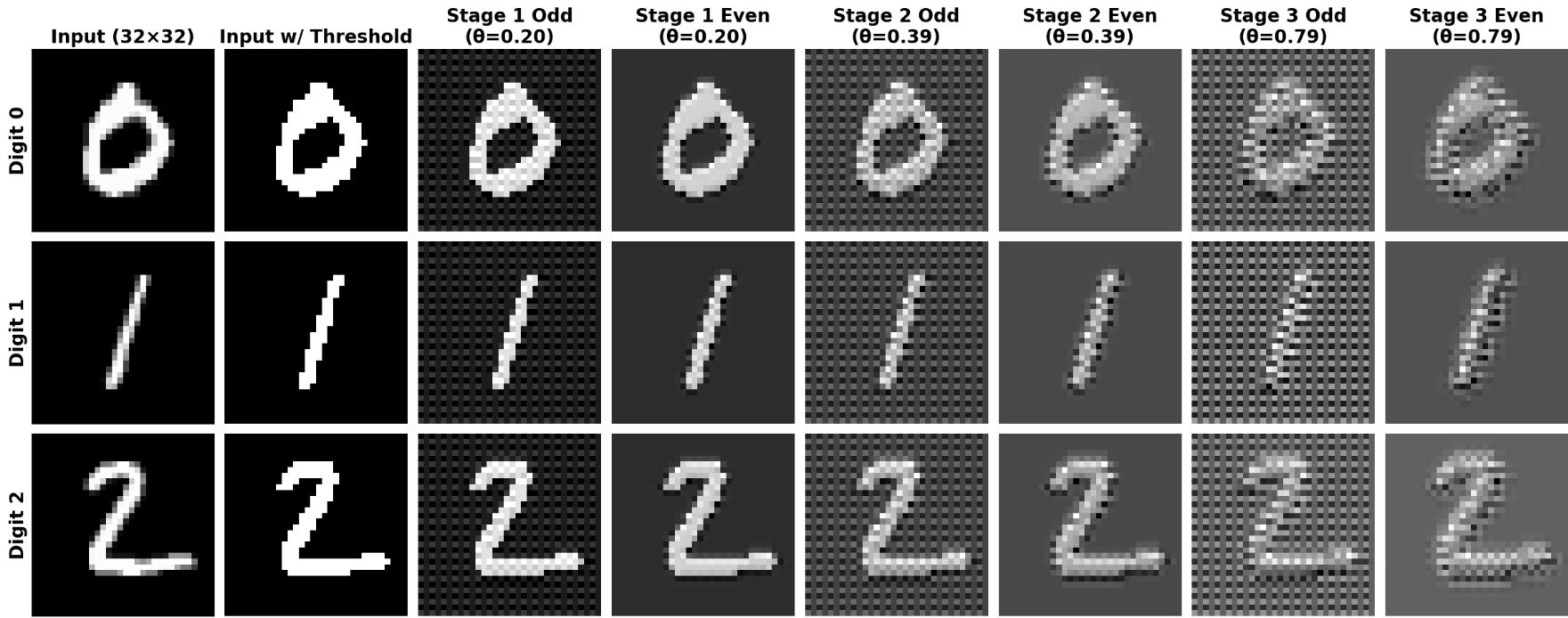} 
    \caption{Unitary convolution of MNIST images for 3 layers, with $\theta$ values as mentioned, after thresholding.}
    \label{fig:ConvolMNIST}
\end{figure}

\begin{figure}[ht]
    \centering
    \includegraphics[width=\linewidth]{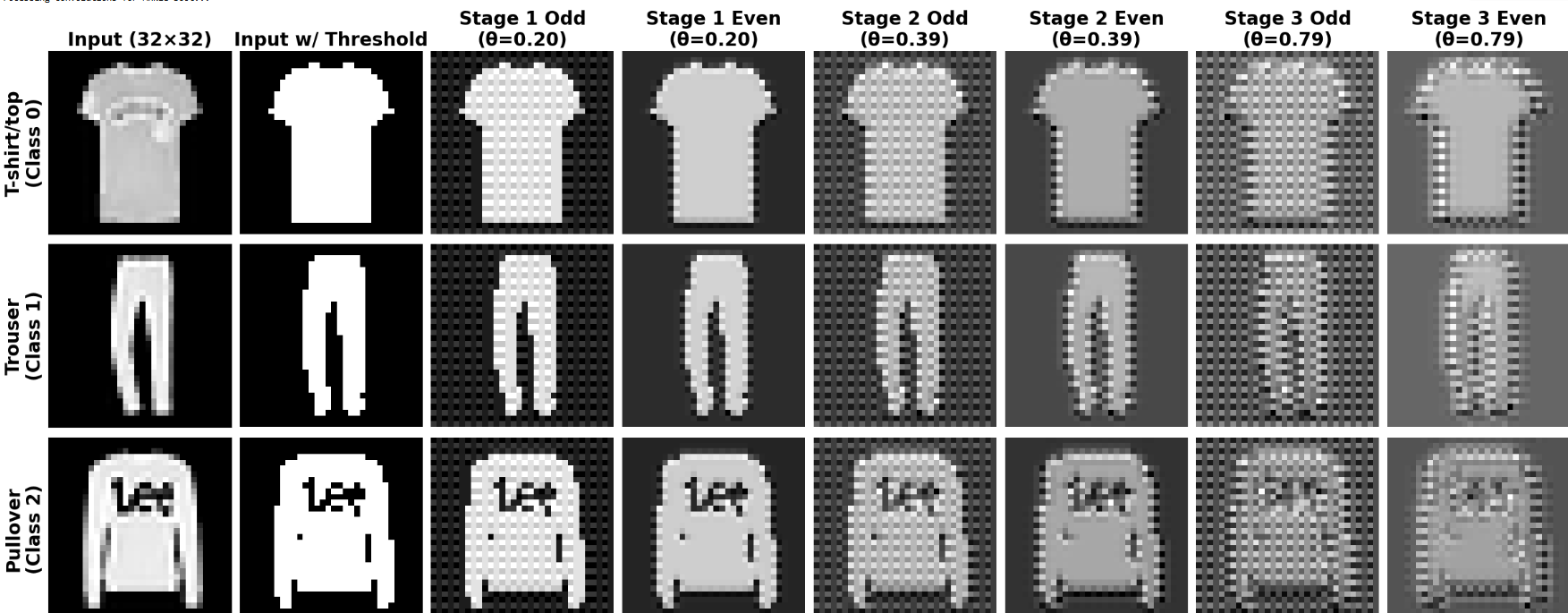} 
    \caption{Unitary convolution of Fashion-MNIST images for 3 layers, with $\theta$ values as mentioned, after thresholding.}
    \label{fig:ConvolFMNIST}
\end{figure}

We show our results for unitary smearing of sample images from MNIST and Fashion-MNIST data sets, in Figs. \ref{fig:ConvolMNIST} and \ref{fig:ConvolFMNIST} respectively, illustrating the effect of thresholding followed by 3 layers of alternate odd and even subset evolution.

\subsection{Multiple Experts}
We identify the quantum experts by the parameters $\theta$ they use for the convolutional smearing, whereby they transform the image features differently. For $N_{exp}$ experts, their individual parameters $\theta_i$ are clamped in the intervals $[(i-1)\pi/N_{exp},i\pi/N_{exp}]$ during the optimisation by FCL, so as to keep them distinct. Each expert provides 16 syndrome features that are passed on to the FCL. In our quantum-inspired setting, the feature for syndrome generator $M$ is easily calculated as the expectation value $\langle\psi|M|\psi\rangle$ using vectorised operations. For independent experts, the classification prediction is made based on their individual 16 features. On the other hand, for the joint expert scenario, the classification prediction is made based on their combined 16$N_{exp}$ features.

\subsection{Classification}
For maximal hardware utilization, we employed \texttt{torch.set\_float32\_matmul\_precision('high')} to enforce TensorFloat-32 (TF32) matrix multiplications on NVIDIA GPUs. Also, to reduce framework overhead, the quantum model was compiled into optimized C++ kernels via \texttt{torch.compile}. The FCL implemented Leaky ReLU activations, and batch normalization (with batch size 256) was performed across the layers. The model incorporated a dropout rate of 0.2 and was trained for 30 epochs. The learning rate began at 0.002, and was regulated by a scheduler with a decay factor of 0.5 and a patience parameter of 1. The CrossEntropyLoss label smoothing parameter was set to 0.05.

To benchmark the efficacy of our proposed quantum-inspired algorithm, we adopted the results obtained with classical diffusion (with varying diffusion parameter $\alpha$) and block-wise pooling as the baseline. The comparison then focused on class prediction accuracy and the number of parameters involved in the analysis, while still maintaining the algorithm's capability to preserve the underlying features of interest. The complexity of the algorithm was dictated by two core parameters: the number of quantum experts and the number of quantum convolution layers.

In the independent expert case, the feature tensor produced by pooling was sliced, completely separating the analysis of individual experts. The cross entropy loss was computed independently for every expert, penalising each isolated pathway for its specific domain errors. For class prediction, the network collected the results of all the decoupled experts and made the decision by majority vote.

In the joint expert case, all the features produced by pooling were processed together in the cross entropy loss computation. We found that this strategy consistently outperformed the independent expert case in class prediction. Table \ref{tab:full_comparison} exemplifies our results, contrasting the joint expert case results with those of the individual expert case as well as the corresponding classical baseline. The improved performance of the joint expert algorithm over the classical baseline is consistent with our expectation that deterministic quantum convolution would retain full image information compared to the lossy smearing of classical diffusion. Moreover, the time per epoch for the joint expert case shows only a moderate increase compared to the classical case, much less than the values of $N_{exp}$ and $N_{lyr}$ (both are effectively 1 in the classical case), making the joint expert algorithm practical. We therefore concentrated on optimising the joint expert algorithm to obtain the best possible classification performance.
    
\begin{table*}[ht]
    \centering
    \caption{Performance comparison between quantum and classical classification algorithms. Here $N_{exp}$ is the number of experts, $N_{lyr}$ is the number of convolution layers, Acc the accuracy for joint expert prediction, I-Acc is the accuracy for independent expert prediction, and $T_E$ the time per epoch in seconds (without the initial overhead for loading data).}
    \label{tab:full_comparison}
    \begin{tabular}{@{} ccccc ccc | ccccc ccc @{}}
    \toprule
    
    \multicolumn{9}{c}{\textbf{MNIST}} & \multicolumn{7}{c}{\textbf{Fashion-MNIST}} \\
    \cmidrule(r){1-8} \cmidrule(l){9-16} 
    
    \multicolumn{5}{c}{Quantum} & \multicolumn{3}{c|}{Classical} & \multicolumn{5}{c}{Quantum} & \multicolumn{3}{c}{Classical} \\
    \cmidrule(r){1-5} \cmidrule(lr){6-8} \cmidrule(lr){9-13} \cmidrule(l){14-16} 
    $N_{exp}$ & $N_{lyr}$ & $T_E$(s) & Acc(\%) & I-Acc(\%) & $\alpha$ & $T_E$(s) & Acc(\%) & $N_{exp}$ & $N_{lyr}$ & $T_E$(s) & Acc(\%) & I-Acc(\%) & $\alpha$ & $T_E$(s) & Acc(\%) \\
    \midrule
    
    32 & 14 &4.00& 97.94 & 90.00 & 1.5 & 1.35 & 95.02 & 16 & 32 &5.08& 86.04 & 81.01 & 1.5 & 1.2 & 76.80 \\
    14 & 12 &2.66& 97.73 & 91.23 & 1.0 & 1.44 & 94.62 & 16 & 16 &3.06& 85.88 & 81.52 & 1.0 & 1.3 & 76.30 \\
    12 & 10 &2.39& 97.27 & 91.93 & 0.5 & 1.17 & 93.27 & 14 & 14 &2.66& 85.81 & 81.16 & 0.5 & 1.3 & 75.90 \\
    
    \bottomrule
    \end{tabular}
\end{table*}

\subsection{Mixture of Experts: Detailed Joint Classification Results}    
We systematically varied the values of $N_{exp}$ and $N_{lyr}$ to find the combination that yields the best classification accuracy. By studying the training and test accuracy distributions, we observed that increasing $N_{exp}$ and $N_{lyr}$ initially gives a sharp increase in the classification accuracy, which suggests that a finer discretisation of the parameter space helps in capturing a richer set of image properties. But increasing $N_{exp}$ and $N_{lyr}$ beyond a point produces diminishing returns with increasing overfitting, which is also consistent with the general empirical FCL behaviour. To better understand how the image features affect the class prediction, we also looked at the confusion matrices for the best parameter values.

\subsubsection*{1. MNIST Dataset}
\begin{figure}[ht]
    \centering
    \begin{minipage}{0.48\textwidth}
        \centering
        \includegraphics[width=\linewidth]{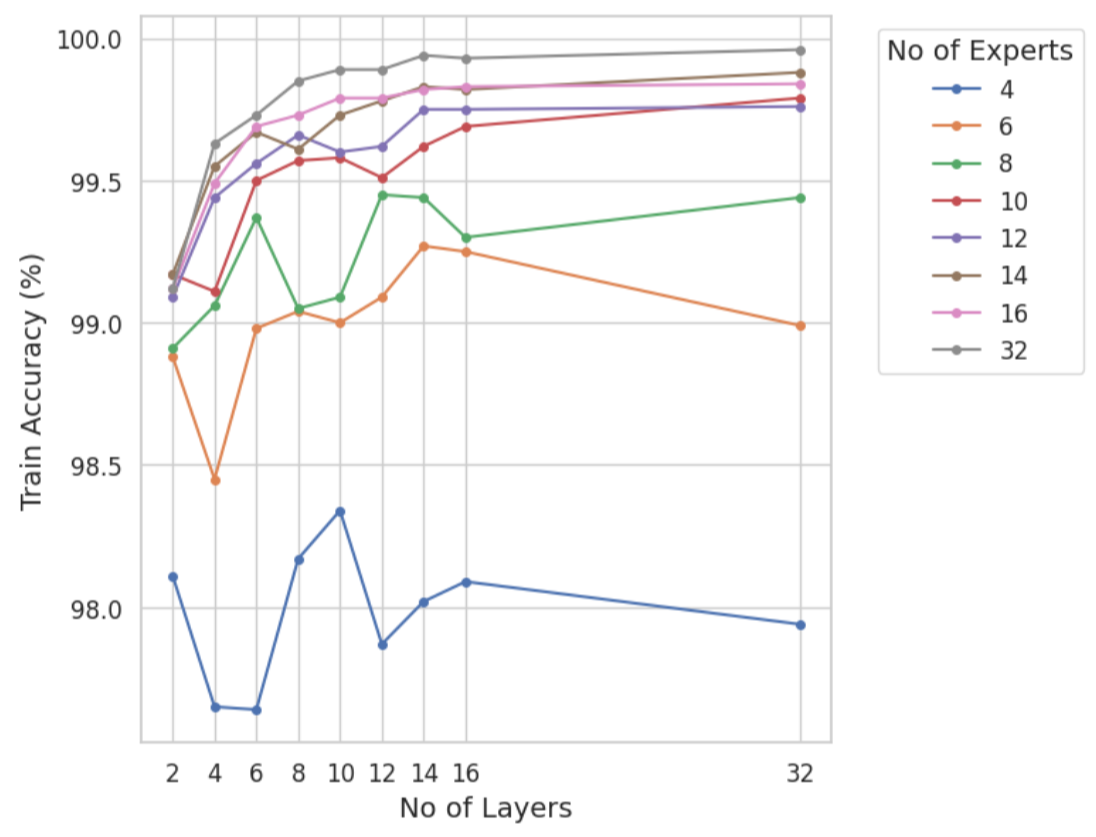}
        \caption{Quantum joint expert performance for the training MNIST dataset, with respect to $N_{exp}$ and $N_{lyr}$.}
        \label{fig:circuit_left4}
    \end{minipage}\hfill
    \begin{minipage}{0.48\textwidth}
        \centering
        \includegraphics[width=\linewidth]{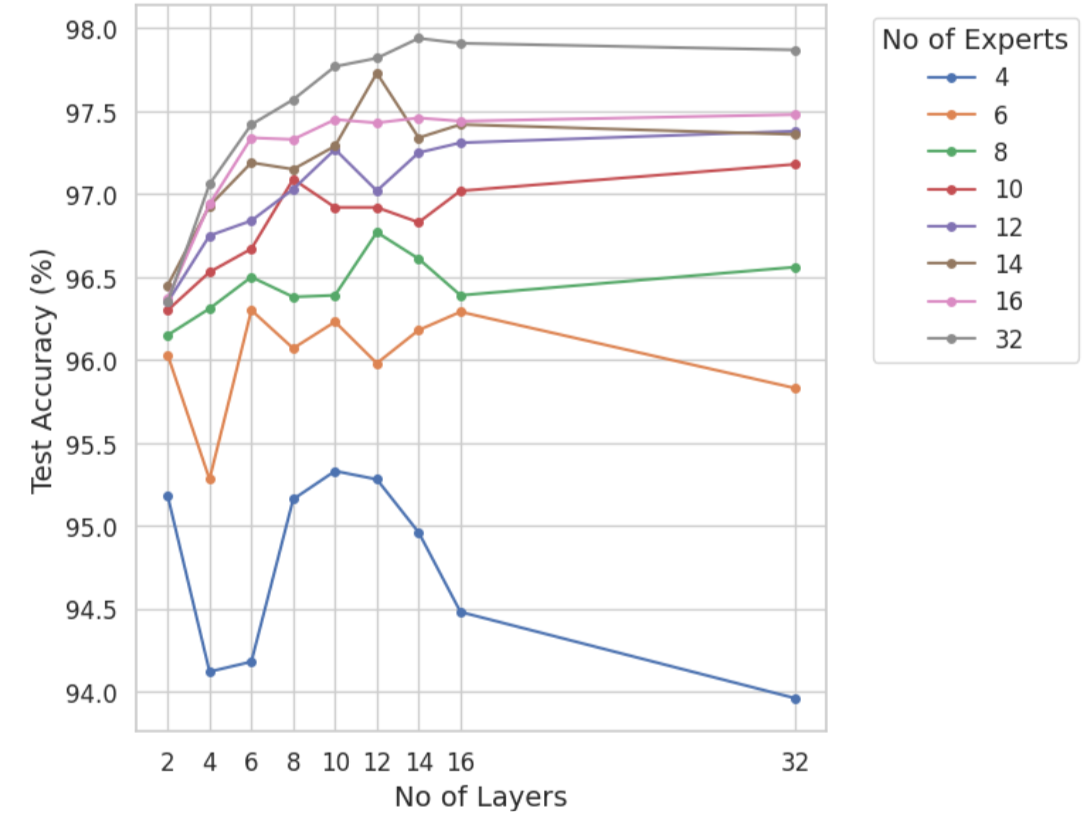}
        \caption{Quantum joint expert performance for the test MNIST dataset, with respect to $N_{exp}$ and $N_{lyr}$.}
        \label{fig:circuit_right4}
    \end{minipage}
\end{figure}

\begin{figure}[ht]
    \centering
    \begin{minipage}{0.48\textwidth}
        \centering
        \includegraphics[width=1.15\linewidth]{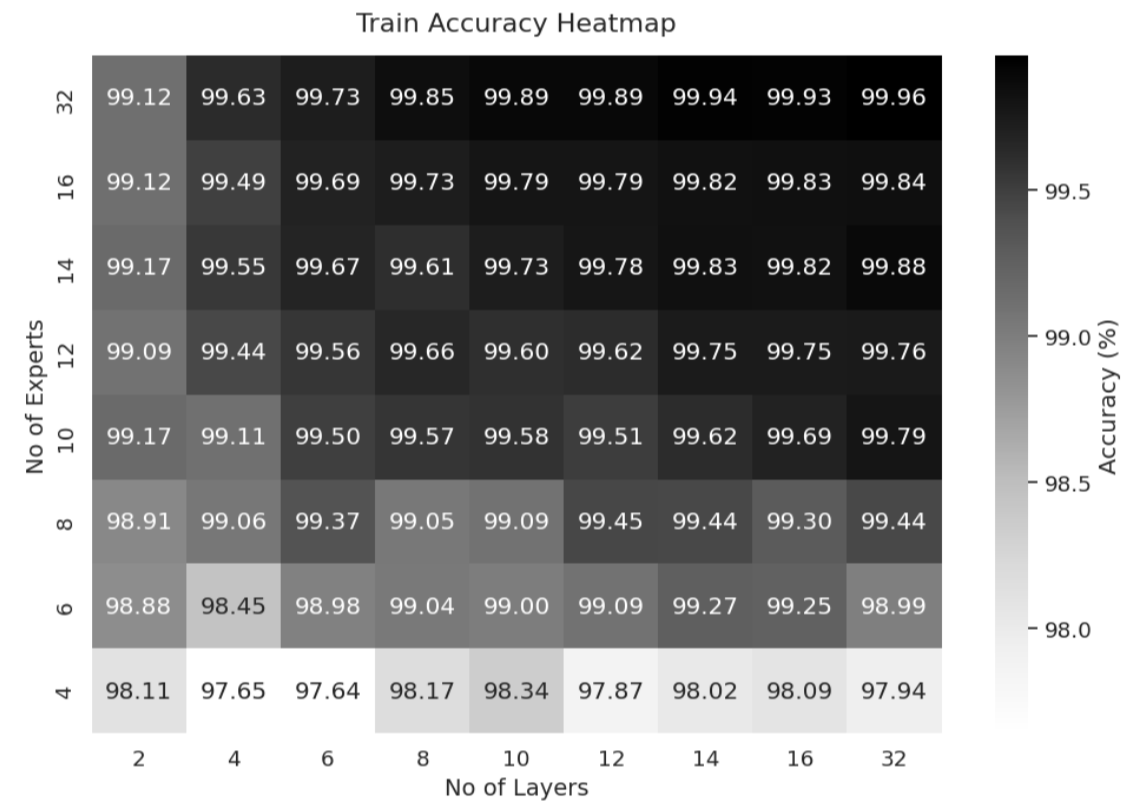}
        \caption{Heatmap for quantum joint expert performance for the training MNIST dataset}
        \label{fig:circuit_left5}
    \end{minipage}\hfill
    \begin{minipage}{0.48\textwidth}
        \centering
        \includegraphics[width=1.15\linewidth]{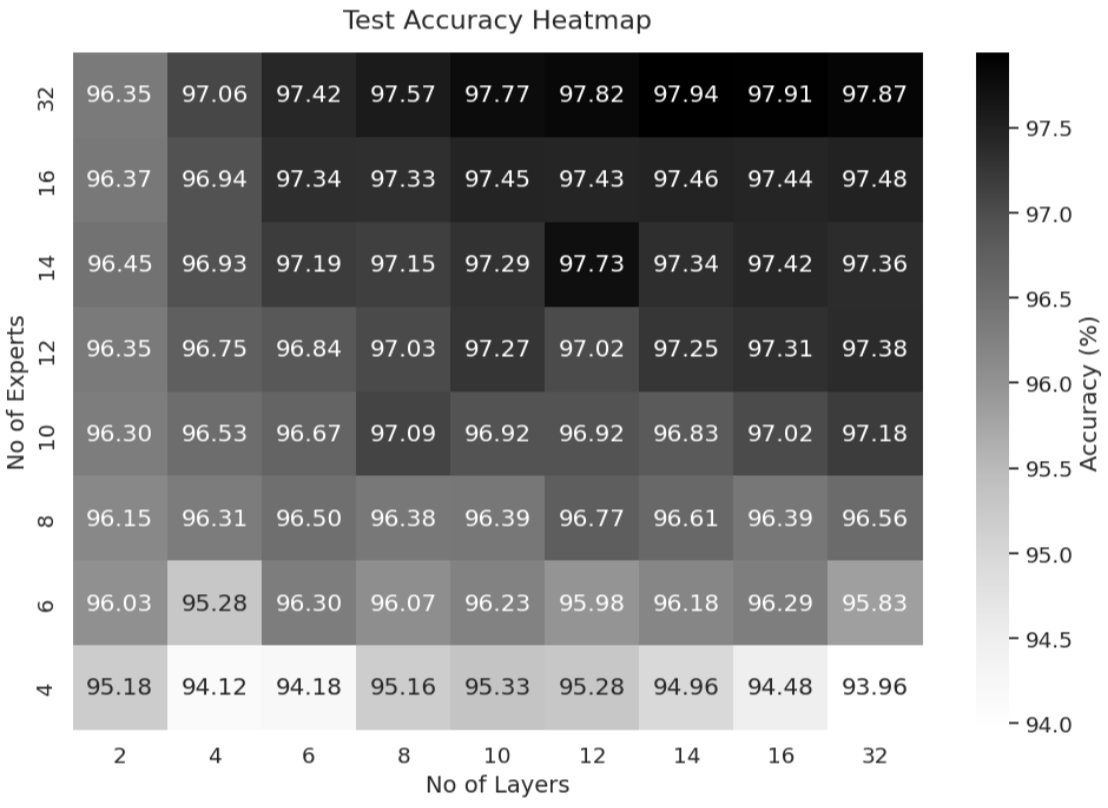}
        \caption{Heatmap for quantum joint expert performance for the test MNIST dataset}
        \label{fig:circuit_right5}
    \end{minipage}
\end{figure}

Our classification accuracy results for the training and test datasets, while varying $N_{exp}$ and $N_{lyr}$, are presented in Figs. \ref{fig:circuit_left4} and \ref{fig:circuit_right4} in graphical form and in Figs. \ref{fig:circuit_left5} and \ref{fig:circuit_right5} as heatmaps. Going up from small values of $N_{exp}$ and $N_{lyr}$, the classification accuracy initially increases substantially but then levels off. Giving more emphasis to the test dataset accuracy, we identified the robust optimal performance point as $N_{exp}$=14 paired with $N_{lyr}$=12. In this configuration, the algorithm yields a test accuracy of 97.73\% and a training accuracy of 99.78\%. We also noticed that the accuracy reached close to its final value by about 20 epochs, indicating that the algorithm converges well. The classification accuracy increases slightly with still higher values of $N_{exp}$ and $N_{lyr}$, but that is a matter of diminishing returns. Our quantum-inspired classification algorithm thus attains good accuracy at moderate depths, while more than halving the failure rate compared to the classical baseline, which should make it a highly viable choice in practice.

\begin{figure}[ht]
    \centering
    \includegraphics[width=0.6\linewidth]{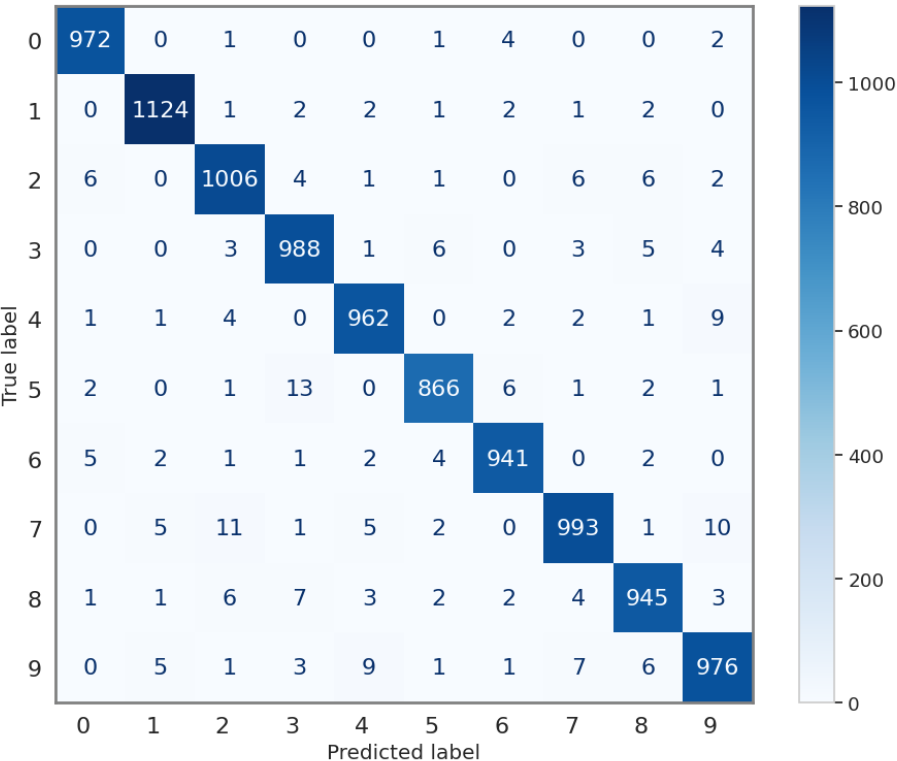} 
    \caption{Confusion matrix for the optimal quantum joint expert analysis of the MNIST test dataset}
    \label{fig:test_accuracy7}
\end{figure}

The confusion matrix for our optimal choice of mixture of experts algorithm is shown in Fig. \ref{fig:test_accuracy7}. It shows excellent class separation among the digits of the MNIST test dataset. The misidentifications are rather small; prominent among them are those between digits 2$\leftrightarrow$7, 3$\leftrightarrow$5, 4$\leftrightarrow$9 and 7$\leftrightarrow$9, which can all be reasonably guessed from the handwriting style.

\subsubsection*{2. Fashion-MNIST Dataset}
\begin{figure}[ht]
    \centering
    \begin{minipage}{0.48\textwidth}
        \centering
        \includegraphics[width=\linewidth]{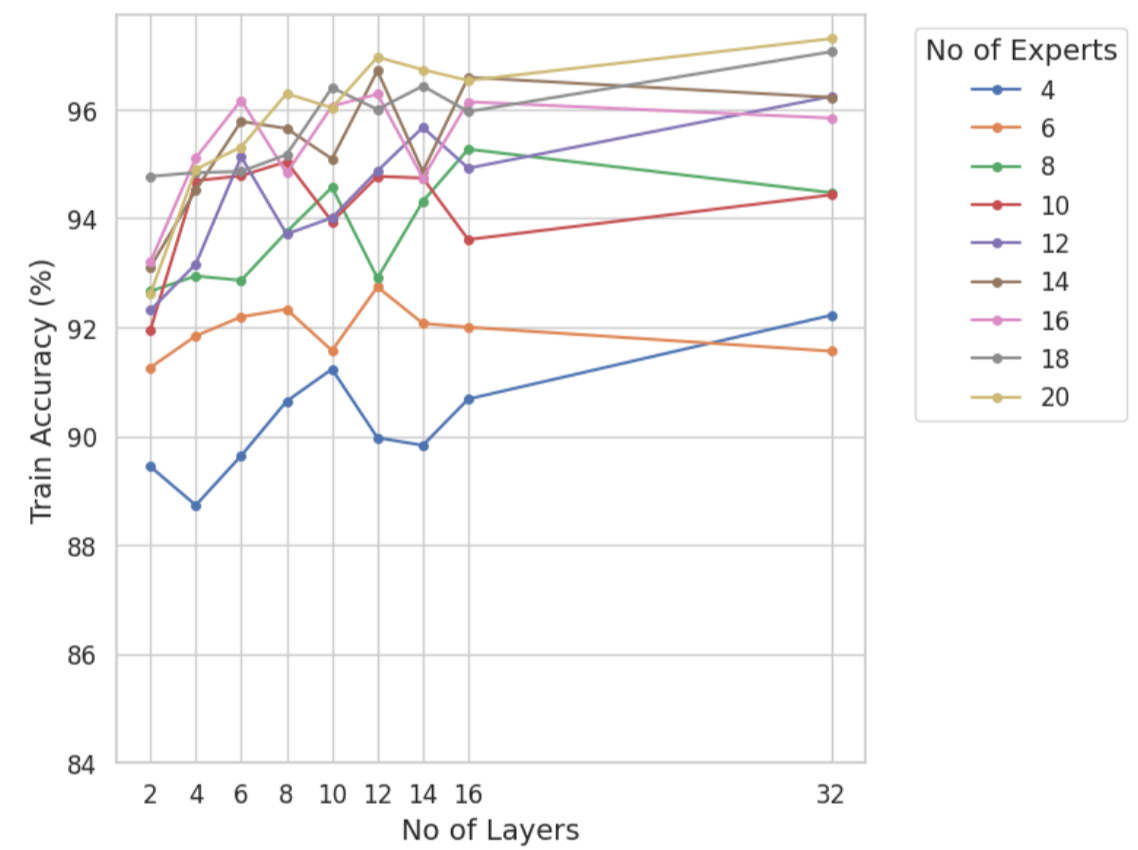}
        \caption{Quantum joint expert performance for the training Fashion-MNIST dataset, with respect to $N_{exp}$ and $N_{lyr}$.}
        \label{fig:circuit_left6}
    \end{minipage}\hfill
    \begin{minipage}{0.48\textwidth}
        \centering
        \includegraphics[width=\linewidth]{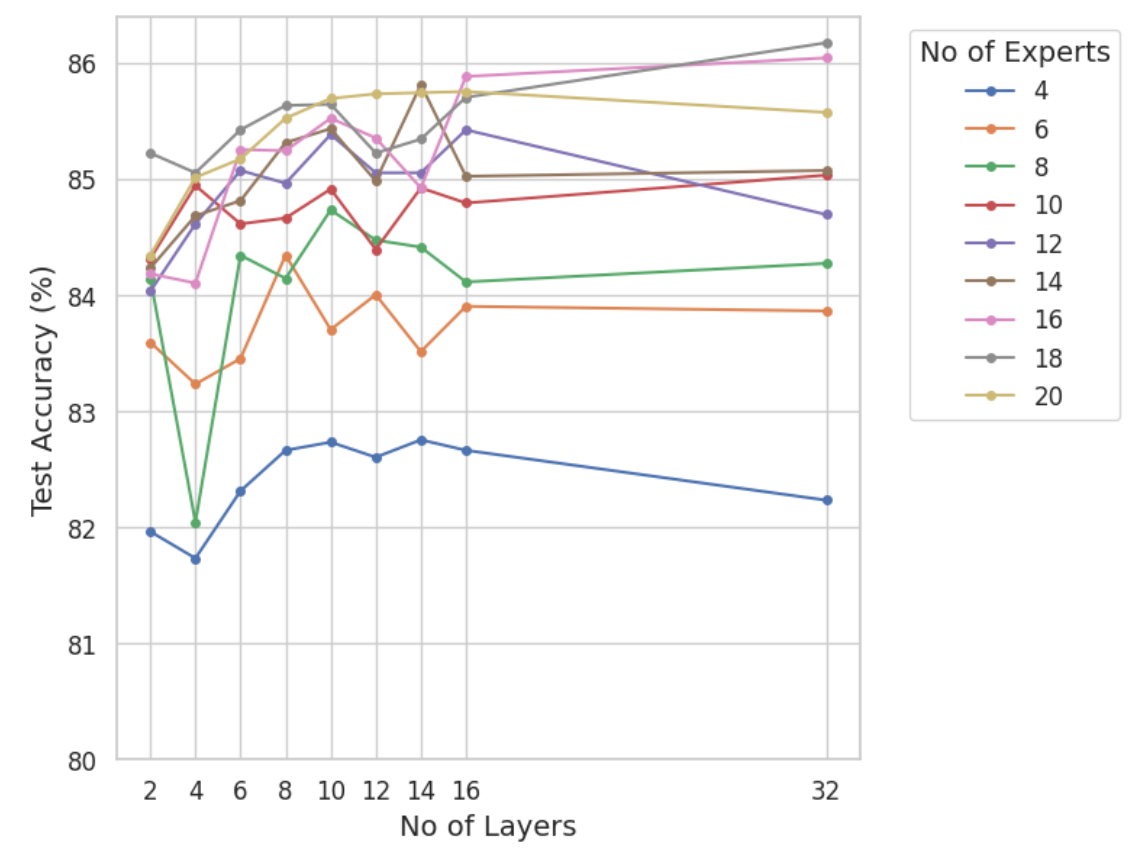}
        \caption{Quantum joint expert performance for the test Fashion MNIST dataset, with respect to $N_{exp}$ and $N_{lyr}$.}
        \label{fig:circuit_right6}
    \end{minipage}
\end{figure}

\begin{figure}[ht]
    \centering
    \begin{minipage}{0.48\textwidth}
        \centering
        \includegraphics[width=1.15\linewidth]{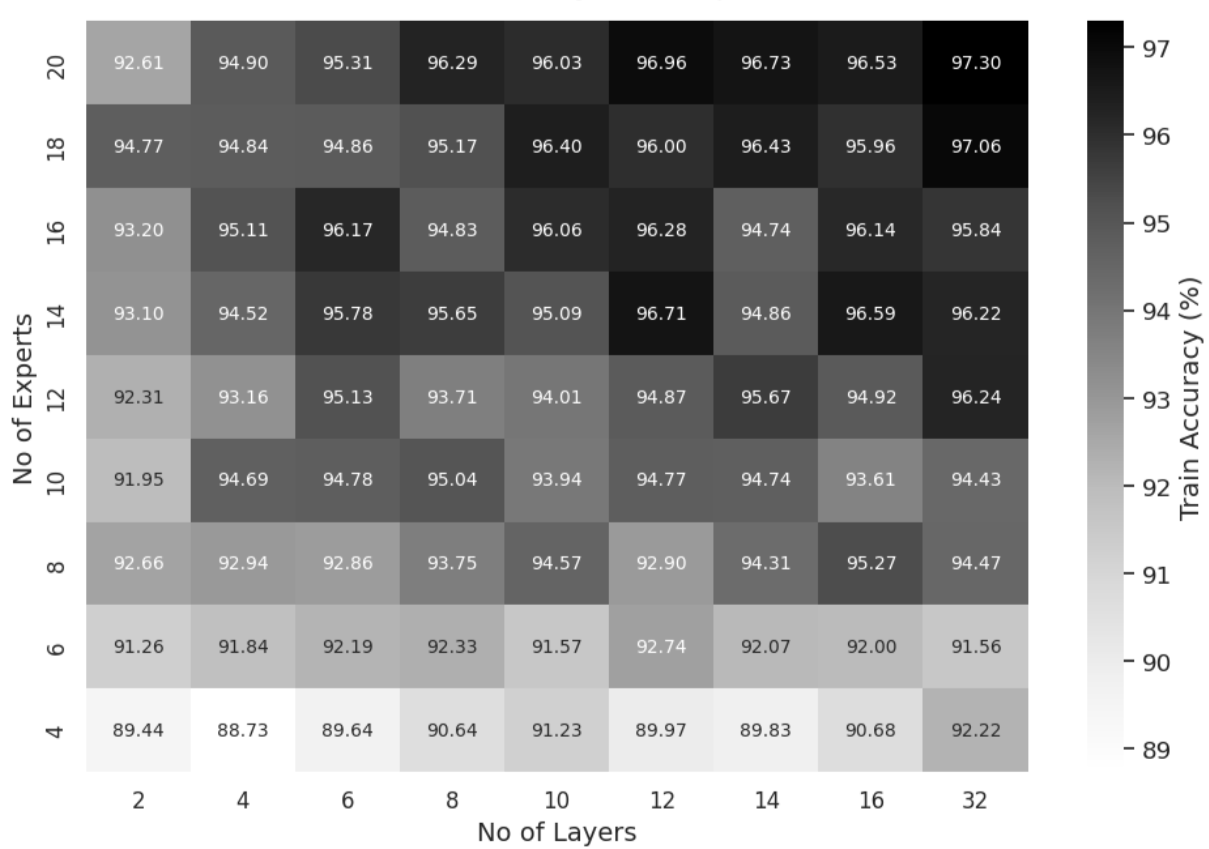}
        \caption{Heatmap for quantum joint expert performance for the training Fashion-MNIST dataset}
        \label{fig:circuit_left7}
    \end{minipage}\hfill
    \begin{minipage}{0.48\textwidth}
        \centering
        \includegraphics[width=1.15\linewidth]{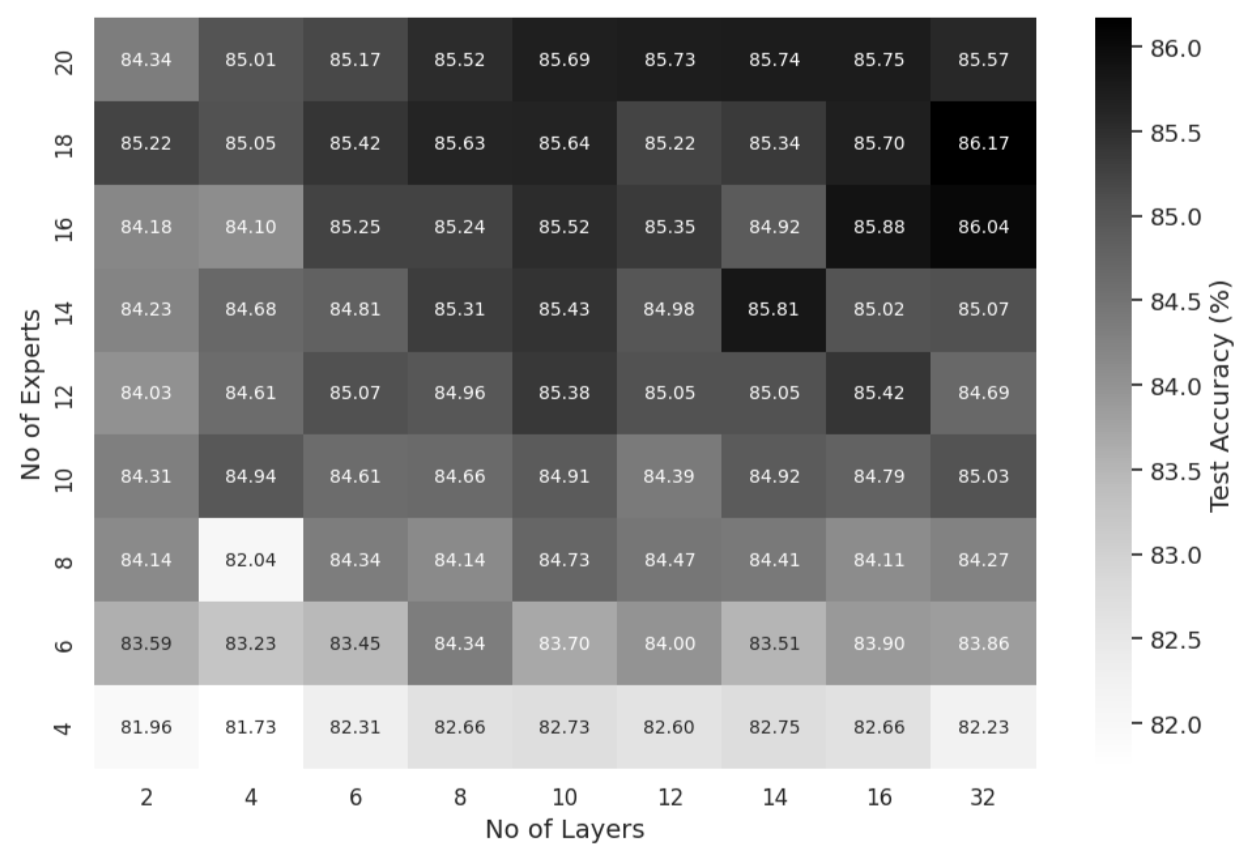}
        \caption{Heatmap for quantum joint expert performance for the test Fashion-MNIST dataset}
        \label{fig:circuit_right7}
    \end{minipage}
\end{figure}

Our classification accuracy results for the training and test datasets, while varying $N_{exp}$ and $N_{lyr}$, are presented in Figs. \ref{fig:circuit_left6} and \ref{fig:circuit_right6} in graphical form and in Figs. \ref{fig:circuit_left7} and \ref{fig:circuit_right7} as heatmaps. The Fashion-MNIST dataset has more complex features than the MNIST dataset, with somewhat ambiguous shapes of spatial boundaries, and so provides a more stringent test of our classification algorithm.

The higher classification difficulty of the Fashion-MNIST dataset is also reflected in the result that the class prediction accuracy with classical diffusion peaks at around 76\%. Our multi-expert quantum-inspired algorithm performs substantially better, bringing down the failure rate by about a factor of two. We identified the optimal performance point as $N_{exp}=16$ and $N_{lyr}=16$, although $N_{exp}=14$ and $N_{lyr}=14$ is almost as good. There the test accuracy is 85.88\% and the training accuracy is 96.14\%. The overall qualitative trends of the classification accuracy are similar to those for the MNIST dataset.

\begin{figure}[ht]
    \centering
    \includegraphics[width=0.6\linewidth]{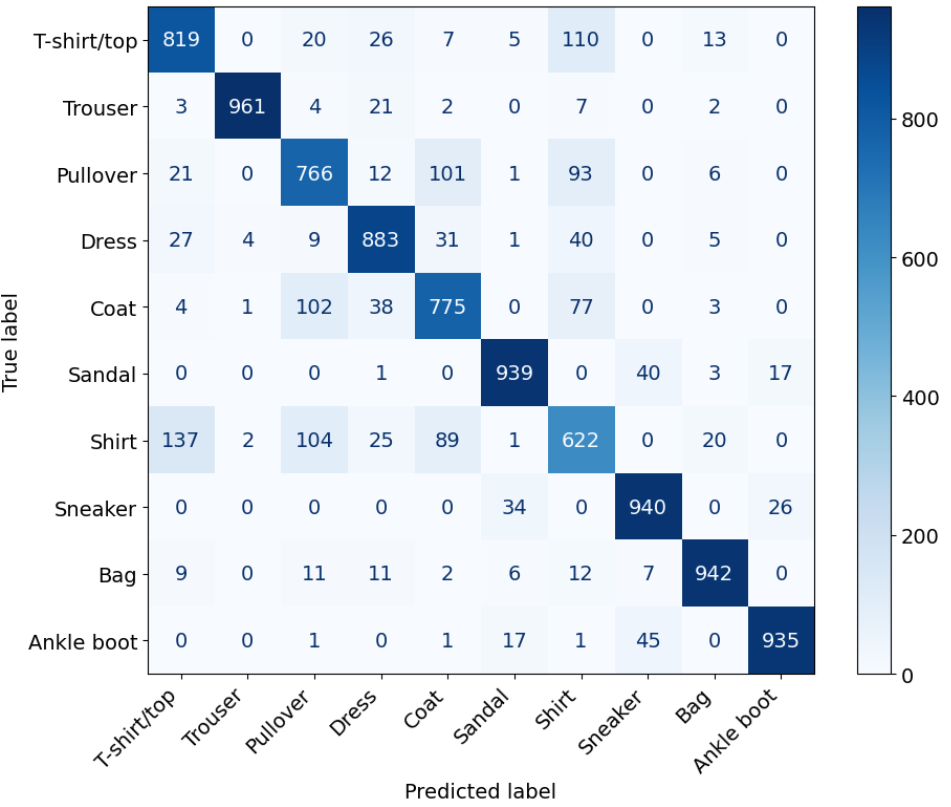} 
    \caption{Confusion matrix for the optimal quantum joint expert analysis of Fashion-MNIST test dataset}
    \label{fig:test_accuracy8}
\end{figure}

The confusion matrix for our optimal choice of mixture of experts algorithm is shown in Fig. \ref{fig:test_accuracy8}. It shows very good separation among the garment shapes of the Fashion-MNIST test dataset.
The misclassifications highlight similarity of shapes among the garments, such as Shirt vs. T-shirt/top, Shirt vs. Pullover, Coat vs. Pullover, and Shirt vs. Coat. These specifics reveal the limitations of our feature extraction method. Garments like shirts, T-shirt/tops, coats and pullovers share similar silhouettes (e.g., long sleeves and common torso coverage), which lead to similar pooling syndromes. Note that classes with distinct shapes (e.g., Trouser, Bag and Ankle boot) have little trouble getting differentiated. 

\section{Conclusions}

In this work, our objective has been to devise a strategy for extracting as much information as possible from the input data, while giving up the details to reduce the computational effort, in the CNN framework for classifying images. We formulated a hybrid classical-quantum strategy for this purpose, and have illustrated its performance on the standard MNIST and Fashion-MNIST datasets. Specifically,
\begin{itemize}
    \item We converted the images by amplitude encoding, from 28$\times$28 pixels to 5$\times$5 qubit states. This reduction of the image size exponentially brought down the complexity of subsequent operations.
    \item We replaced the classical diffusion-based lossy convolution by quantum local unitary convolution with even-odd partitioning of the image grid. In the process, we introduced multiple experts performing convolution with different smearing parameters.
    \item For feature extraction from the images, we replaced the classical block-wise pooling method by the syndromes of a quantum stabiliser code.
    \item We then analysed the extracted 16$N_{exp}$ features by the experts, jointly using the conventional FCL framework. This number of features is orders of magnitude smaller than the number of parameters needed in the FCL analysis of the original 28$\times$28 pixel size images.
\end{itemize}

Our results rigorously demonstrate that our quantum-inspired strategy is superior to the conventional classical method for the feature extraction and classification tasks. This is noteworthy for the MNIST dataset, and much more so for the Fashion-MNIST dataset with higher shape ambiguity. In particular, 
\begin{itemize}
    \item The class prediction accuracy by the joint expert analysis clearly outperforms that by the independent expert analysis, implying that our mixture of experts strategy extracts more of the relevant features from the images.
    \item The class prediction failure rate for our joint expert analysis is roughly half of that for the conventional classical method, indicating that our unitary convolution and stabiliser code based syndrome extraction cause less information degradation than the feature extraction methods based on classical diffusion.
    Moreover, the computational overhead introduced by our replacement is quite moderate and tolerable.
\end{itemize}
We have thus firmly established that our quantum-inspired mixture of expert strategy offers a clear advantage in image classification tasks. How it can be implemented in a hybrid classical-quantum processor architecture is described in the Appendix.

Future research on the topic can address how to extend our strategy to more structurally complex diagnostic datasets. For example, anomalies can be identified for early disease detection in medical imaging (such as chest radiography or retinal scans), as well as agricultural and environmental pathologies can be investigated by complex pattern recognition in images. Needless to say, the strategy of performance optimisation using a mixture of experts can be applied to other machine learning tasks, such as reinforcement learning and unsupervised learning. Even physical implementation on neuromorphic computing architectures can be attempted, with promising advantage for latency and power consumption. And, no doubt, the strategy will become part of multi-agent AI models.

\section*{Acknowledgements}
This work was supported by the project ``Quantum Algorithms and Simulations" under the DIA-RCoE at IISc Bangalore.
We thank R.K. Ramakrishnan for his comments on the manuscript.


\section*{Appendix}

Here we describe how various quantum-inspired parts of our algorithm can be implemented on a quantum processor.

\subsection{Amplitude Encoding}
We have opted for amplitude encoding of the quantum state after thresholding. That fixes the normalisation of the state specified in Eq. \ref{eq:amp_encoding}. Such a state can be generated in a straightforward manner, by starting from $|00000,00000\rangle$ and applying rotations by fixed angles while sequentially going through the pixels labeled by $x,y\in{0,1}^5$. Each new rotation adds to the state a component orthogonal to the ones already included, which uniquely specifies the rotation direction while the value of $c_{x,y}$ specifies the rotation angle.

Amplitude encoding efficiently encodes the quantum state in a small number of qubits, but then operations on the state require a higher number of logic gates. We need to keep in mind that implementation of the quantum algorithm on the currently available noisy quantum processors requires careful management of both space and time resources, to ensure efficiency as well as reliability.

An alternative is angle encoding, which maps each classical pixel information (i.e. the $c_{x,y}$ value before thresholding) to the rotation angle of one qubit about a chosen axis. That leads to shallow logic circuits, but the number of qubits remains large. We find that unsuitable for our quantum algorithm, and so do not consider the possibility of angle encoding.

\subsection{Quantum Convolution}
The 4$\times$4 block matrices of our unitary convolution, Eq. \ref{eq:unitary_blocks}, operate on elementary squares
whose vertices have labels \{00,10,01,11\}. Hence, they are 2-qubit logical operations. In the spirit of Euler angle decomposition, they can be factored in terms of 2$\times$2 submatrices as
\begin{equation}
    U_o = \begin{pmatrix} I & 0 \\ 0 & U_2 \end{pmatrix} \begin{pmatrix} CI & SI \\ -SI & CI \end{pmatrix} \begin{pmatrix} U_3 & 0 \\ 0 & \sigma_3 \end{pmatrix} ,
\end{equation}
where $C=\sqrt{c^2+s'^2}$, $S=s'=s/\sqrt{2}$ and $C^2+S^2=1=c^2+s^2$, together with
\begin{equation}
    U_3 = \frac{1}{\sqrt{c^2 + s'^2}} \begin{pmatrix} c & s' \\ -s' & c \end{pmatrix} ,~~ U_2 = U_3 \sigma_3 .
\end{equation}
Note that our convention has kept all components real, completely avoiding complex numbers.

\begin{figure}[ht]
\begin{center}
{\setlength{\unitlength}{1mm}
\begin{picture}(85,30)
\thicklines
\put(6,25){\line(1,0){30}}
\put(48,25){\line(1,0){30}}
\put(80,24){$q_0$}
\put(6,5){\line(1,0){4}}
\put(16,5){\line(1,0){4}}
\put(32,5){\line(1,0){4}}
\put(48,5){\line(1,0){4}}
\put(58,5){\line(1,0){4}}
\put(74,5){\line(1,0){4}}
\put(80,4){$q_1$}
\put(10,2){\line(1,0){6}}
\put(10,2){\line(0,1){6}}
\put(16,2){\line(0,1){6}}
\put(10,8){\line(1,0){6}}
\put(13,8){\line(0,1){17}}
\put(13,25){\circle*{1.5}}
\put(11.5,4){$Z$}
\put(20,2){\line(1,0){12}}
\put(20,2){\line(0,1){6}}
\put(32,2){\line(0,1){6}}
\put(20,8){\line(1,0){12}}
\put(26,8){\line(0,1){17}}
\put(26,25){\circle*{1.5}}
\put(20.5,4){$R_y(-\alpha)$}
\put(36,2){\line(1,0){12}}
\put(36,2){\line(0,1){6}}
\put(48,2){\line(0,1){6}}
\put(36,8){\line(1,0){12}}
\put(37,4){$R_y(\alpha)$}
\put(36,22){\line(1,0){12}}
\put(36,22){\line(0,1){6}}
\put(48,22){\line(0,1){6}}
\put(36,28){\line(1,0){12}}
\put(37,24){$R_y(\beta)$}
\put(52,2){\line(1,0){6}}
\put(52,2){\line(0,1){6}}
\put(58,2){\line(0,1){6}}
\put(52,8){\line(1,0){6}}
\put(55,8){\line(0,1){17}}
\put(55,25){\circle*{1.5}}
\put(53.5,4){$Z$}
\put(62,2){\line(1,0){12}}
\put(62,2){\line(0,1){6}}
\put(74,2){\line(0,1){6}}
\put(62,8){\line(1,0){12}}
\put(68,8){\line(0,1){17}}
\put(68,25){\circle*{1.5}}
\put(63,4){$R_y(\alpha)$}
\end{picture}
}
\end{center}
\caption{Two-qubit logic gate circuit for the operator $U_o$}
\label{fig:Circuit_Uo}
\end{figure}
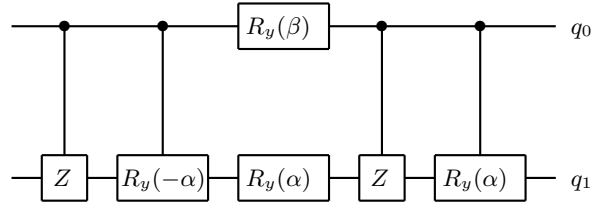

We convert this decomposition into elementary quantum logic gates, by rewriting it as
\begin{equation}
    U_o = \begin{pmatrix} I & 0 \\ 0 & U_2 \end{pmatrix}
    \left[ \begin{pmatrix} C & S \\ -S & C \end{pmatrix} \otimes I \right]
    \left[ \begin{pmatrix} U_3 & 0 \\ 0 & U_3 \end{pmatrix} \begin{pmatrix} I & 0 \\ 0 & U_3^\dagger \sigma_3 \end{pmatrix} \right] ,
\end{equation}
and the resultant quantum logic gate circuit for $U_o$ is shown in Fig. \ref{fig:Circuit_Uo}.

\noindent Here, the rotation operator is $R_y(\phi)=\exp(i\phi\sigma_2)$, and the rotation angles are $\alpha=\tan^{-1}(s'/c)$ and $\beta=\tan^{-1}(S/C)$.

\subsection{Coordinate Translation}
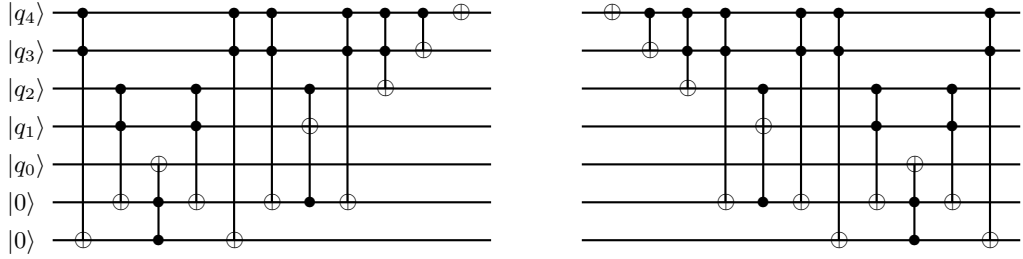
\begin{figure}[ht]
\begin{center}
{\setlength{\unitlength}{1mm}
\begin{picture}(140,40)
\thicklines
\put(0,34){$|q_4\rangle$} \put(6,35){\line(1,0){58}}
\put(0,29){$|q_3\rangle$} \put(6,30){\line(1,0){58}}
\put(0,24){$|q_2\rangle$} \put(6,25){\line(1,0){58}}
\put(0,19){$|q_1\rangle$} \put(6,20){\line(1,0){58}}
\put(0,14){$|q_0\rangle$} \put(6,15){\line(1,0){58}}
\put(0,9){$|0\rangle$} \put(6,10){\line(1,0){58}}
\put(0,4){$|0\rangle$}  \put(6,5){\line(1,0){58}}
\put(10,5){\line(0,1){30}} \put(10,35){\circle*{1.5}} \put(10,30){\circle*{1.5}} \put(8.8,4.3){$\oplus$}
\put(15,10){\line(0,1){15}} \put(15,25){\circle*{1.5}} \put(15,20){\circle*{1.5}} \put(13.8,9.3){$\oplus$}
\put(20,5){\line(0,1){10}} \put(20,5){\circle*{1.5}} \put(20,10){\circle*{1.5}} \put(18.8,14.3){$\oplus$}
\put(25,10){\line(0,1){15}} \put(25,25){\circle*{1.5}} \put(25,20){\circle*{1.5}} \put(23.8,9.3){$\oplus$}
\put(30,5){\line(0,1){30}} \put(30,35){\circle*{1.5}} \put(30,30){\circle*{1.5}} \put(28.8,4.3){$\oplus$}
\put(35,10){\line(0,1){25}} \put(35,35){\circle*{1.5}} \put(35,30){\circle*{1.5}} \put(33.8,9.3){$\oplus$}
\put(40,10){\line(0,1){15}} \put(40,25){\circle*{1.5}} \put(40,10){\circle*{1.5}} \put(38.8,19.3){$\oplus$}
\put(45,10){\line(0,1){25}} \put(45,35){\circle*{1.5}} \put(45,30){\circle*{1.5}} \put(43.8,9.3){$\oplus$}
\put(50,25){\line(0,1){10}} \put(50,35){\circle*{1.5}} \put(50,30){\circle*{1.5}} \put(48.8,24.3){$\oplus$}
\put(55,30){\line(0,1){5}} \put(55,35){\circle*{1.5}} \put(53.8,29.3){$\oplus$}
\put(58.8,34.3){$\oplus$}

\put(76,35){\line(1,0){58}}
\put(76,30){\line(1,0){58}}
\put(76,25){\line(1,0){58}}
\put(76,20){\line(1,0){58}}
\put(76,15){\line(1,0){58}}
\put(76,10){\line(1,0){58}}
\put(76,5){\line(1,0){58}}
\put(130,5){\line(0,1){30}} \put(130,35){\circle*{1.5}} \put(130,30){\circle*{1.5}} \put(128.8,4.3){$\oplus$}
\put(125,10){\line(0,1){15}} \put(125,25){\circle*{1.5}} \put(125,20){\circle*{1.5}} \put(123.8,9.3){$\oplus$}
\put(120,5){\line(0,1){10}} \put(120,5){\circle*{1.5}} \put(120,10){\circle*{1.5}} \put(118.8,14.3){$\oplus$}
\put(115,10){\line(0,1){15}} \put(115,25){\circle*{1.5}} \put(115,20){\circle*{1.5}} \put(113.8,9.3){$\oplus$}
\put(110,5){\line(0,1){30}} \put(110,35){\circle*{1.5}} \put(110,30){\circle*{1.5}} \put(108.8,4.3){$\oplus$}
\put(105,10){\line(0,1){25}} \put(105,35){\circle*{1.5}} \put(105,30){\circle*{1.5}} \put(103.8,9.3){$\oplus$}
\put(100,10){\line(0,1){15}} \put(100,25){\circle*{1.5}} \put(100,10){\circle*{1.5}} \put(98.8,19.3){$\oplus$}
\put(95,10){\line(0,1){25}} \put(95,35){\circle*{1.5}} \put(95,30){\circle*{1.5}} \put(93.8,9.3){$\oplus$}
\put(90,25){\line(0,1){10}} \put(90,35){\circle*{1.5}} \put(90,30){\circle*{1.5}} \put(88.8,24.3){$\oplus$}
\put(85,30){\line(0,1){5}} \put(85,35){\circle*{1.5}} \put(83.8,29.3){$\oplus$}
\put(78.8,34.3){$\oplus$}
\end{picture}
}
\end{center}
\caption{Reversible logic circuit for adding 1 to (left) and subtracting 1 from (right) a 5-qubit register. They are inverses of each other. The least significant bit is at the top, and two ancilla bits are at the bottom.}
\label{fig:Circuit_add_subtract}
\end{figure}

As described in Section \ref{sec:unitary_convol}, the operation of $U_e$ is the same as that of $U_o$, with the parameter $s$ replaced by $-s$ and the vertex labels of elementary square changed to \{00,-10,0-1,-1-1\}. Because of the negative shifts, the vertex labels cannot be covered by just two bits. A simple way out is to translate the whole image so that the point (0,0) maps to point (1,1), apply the operator $U_e$ using two bit vertex labels, and then translate the image back to its original coordinates. This procedure needs logic circuits that first add one to, and later subtract one from, the $x$ and $y$ pixel labels. The reversible logic circuits that accomplish these addition and subtraction tasks, using Toffoli gates and two ancilla bits, are presented in Fig. \ref{fig:Circuit_add_subtract}.

\subsection{Stabiliser Code Syndrome Extraction}
We use the expectation values of the syndromes of the [[5,1,3]] stabiliser code as image features, along both $x$ and $y$ axes. These syndromes contain only $X$ and $Z$ Pauli operators, and can be extracted using the single qubit Hadamard logic gate and parity measurements \cite{diVincenzo_shor_PRL1996}. On a quantum processor, the syndrome measurements give the eigenvalues (+1 or -1) of the generators. To obtain the image features as the expectation values, the syndrome measurements require many shots till the features reach sufficient accuracy. This effort is a computational overhead, because the convergence towards the expectation value scales as $O(N_{shot}^{-1/2})$.

\subsection{Complete Algorithm Structure}
\begin{figure}[ht]
    \centering
    \includegraphics[width=\textwidth]{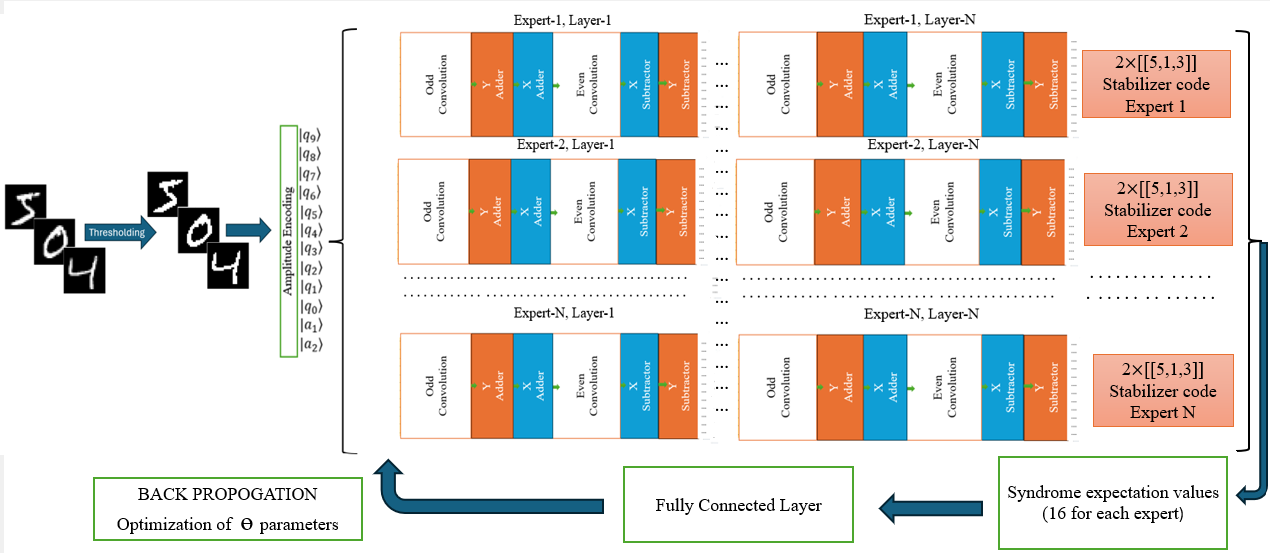} 
    \caption{Block circuit diagram for the full multi-expert image classification algorithm}
    \label{fig:algorithm_block_diagram}
\end{figure}

The complete hybrid classical-quantum structure of our image classification algorithm is presented as a block diagram in Fig. \ref{fig:algorithm_block_diagram}.

\begin{thebibliography}{99}

\bibitem{krizhevsky2012}
A. Krizhevsky, I. Sutskever and G.E. Hinton, ``ImageNet Classification with Deep Convolutional Neural Networks",
Advances Neural Information Processing Systems (NeurIPS) 25 (2012).

\bibitem{hinton2015}
Y. LeCun, Y. Bengio and G. Hinton, ``Deep Learning",
Nature 521, 436-444 (2015).

\bibitem{bermejo2026}
P. Bermejo, P. Braccia, M.S. Rudolph, Z. Holmes, L. Cincio and M. Cerezo, ``Quantum Convolutional Neural Networks are Effectively Classically Simulable",
Phys. Rev. X Quantum 7, 020304 (2026).

\bibitem{bowles2024}
J. Bowles, S. Ahmed and M. Schuld, ``Better than Classical? The Subtle Art of Benchmarking Quantum Machine Learning Models",
[preprint arXiv:2403.07059] (2024).

\bibitem{QMLrev}
R.K. Ramakrishnan, S. Kashyap, K. Jyoti and A.D. Patel, ``Advancing in Machine Learning: Where can Quantum Techniques Help?", [preprint arXiv:2507.08379] (2025).

\bibitem{huang2022}
H.-Y. Huang, M. Broughton, J. Cotler, S. Chen, J. Li, M. Mohseni, H. Neven, R. Babbush, R. Kueng, J. Preskill and J.R. Mcclean, ``Quantum Advantage in Learning from Experiments",
Science 376, 1182-1186 (2022).

\bibitem{MoEref2}
H.-Q. Nguyen et al., ``QMoE: A Quantum Mixture of Experts Framework for Scalable Quantum Neural Networks", 2025 IEEE International Conference on Quantum Computing and Engineering (QCE), Albuquerque, NM, USA, pp. 223-228 [preprint arXiv:2507.05190] (2025).

\bibitem{MoEref1}
P.A.X. Tognini et al., ``Solving MNIST with a Globally Trained Mixture of Quantum Experts",
[preprint arXiv:2505.14789] (2025).

\bibitem{lecun2010mnist}
Y. LeCun, L. Bottou, Y. Bengio and P. Haffner, ``Gradient-based Learning Applied to Document Recognition",
Proc. IEEE 86(11), 2278-2324 (1998).
Image database: http://kaggle.com/datasets/hojjatk/mnist-dataset

\bibitem{xiao2017fashion}
H. Xiao, K. Rasul and R. Vollgraf, ``Fashion-MNIST: A Novel Image Dataset for Benchmarking Machine Learning Algorithms", 
[preprint arXiv:1708.07747] (2017).
Image database: http://kaggle.com/datasets/zalando-research/fashionmnist

\bibitem{patel2005}
A. Patel, K.S. Raghunathan and P. Rungta, ``Quantum Random Walks without a Coin Toss", Proc. Workshop on Quantum Information, Computation and Communication (QICC2005), Kharagpur, India, Allied Publishers, pp. 41-55, [preprint arXiv:0506221 (quant-ph)] (2005).

\bibitem{preskill_chap7}
John Preskill, ``Chapter 7: Quantum Error Correction,'' 
Lecture notes for Ph219/CS219: Quantum Information and Computation, California Institute of Technology, 2020. Available online: https://www.preskill.caltech.edu/ph229/notes/chap7.pdf

\bibitem{kingma2014adam}
D.P. Kingma and J. Ba, ``Adam: A method for stochastic optimization", 
[preprint arXiv:1412.6980] (2014).

\bibitem{diVincenzo_shor_PRL1996}
D.P. DiVincenzo and P.W. Shor, ``Fault-Tolerant Error Correction with Efficient Quantum Codes",
Phys. Rev. Lett. 77, 3260 (1996).
\end{thebibliography}
\end{document}